\documentclass[conference]{IEEEtran}
\IEEEoverridecommandlockouts
\usepackage{times}
\pdfoutput=1
\let\chapter\section
\usepackage[numbers]{natbib}
\usepackage{multicol}
\usepackage[bookmarks=true]{hyperref}

\renewcommand\dbltextfloatsep{3pt}
\renewcommand\dblfloatsep{3pt}
\renewcommand\textfloatsep{3pt}
\usepackage[linesnumbered,ruled]{algorithm2e}
\def\BState{\State\hskip-\ALG@thistlm}
\usepackage{pifont}
\usepackage{booktabs}

\usepackage{multirow}
\usepackage{graphicx}
\usepackage{amsfonts}
\usepackage{amsmath,amssymb,mathrsfs}
\usepackage{array}
\usepackage{flafter}
\usepackage{subfig}
\usepackage[linesnumbered,ruled]{algorithm2e}
\usepackage[format=plain,labelsep=period,font=footnotesize]{caption}
\usepackage[bbgreekl]{mathbbol}
\usepackage{amsfonts}

\DeclareSymbolFontAlphabet{\mathbb}{AMSb}
\DeclareSymbolFontAlphabet{\mathbbl}{bbold}
\usepackage{verbatim}
\usepackage{color}
\usepackage{psfrag}
\usepackage{umoline}
\usepackage{hyperref}
\usepackage{indentfirst}
\newtheorem{theorem}{Theorem}
\newtheorem{problem}{Problem}

\newtheorem{lemma}{Lemma}

\newtheorem{coroll}{Corollary}

\allowdisplaybreaks

\makeindex             
\usepackage[figuresright]{rotating}

\begin{document}

\title{Near-Optimal Belief Space Planning via T-LQG$ ^{\tiny{*}} $}
\author
{
Mohammadhussein Rafieisakhaei$^{1}$, Suman Chakravorty$^{2}$ and P. R. Kumar$^{1}$
\thanks{*This material is based upon work partially supported by NSF under Contract Nos. CNS-1646449 and Science \& Technology Center Grant CCF-0939370, the U.S. Army Research Office under Contract No. W911NF-15-1-0279, and NPRP grant NPRP 8-1531-2-651 from the Qatar National Research Fund, a member of Qatar Foundation.}
\thanks{$^{1}$M. Rafieisakhaei and P. R. Kumar are with the Department of Electrical and Computer Engineering, and $^{2}$S. Chakravorty is with the Department of Aerospace Engineering, Texas A\&M University, College Station, Texas, 77840 USA.
        \{\tt\small mrafieis, schakrav, prk@tamu.edu\}}%
}

\maketitle

\begin{abstract}
We consider the problem of planning under observation and motion uncertainty for nonlinear robotics systems. Determining the optimal solution to this problem, generally formulated as a Partially Observed Markov Decision Process (POMDP), is computationally intractable. We propose a Trajectory-optimized Linear Quadratic Gaussian (T-LQG) approach that leads to quantifiably near-optimal solutions for the POMDP problem. We provide a novel ``separation principle'' for the design of an optimal nominal open-loop trajectory followed by an optimal feedback control law, which provides a near-optimal feedback control policy for belief space planning problems involving a polynomial order of calculations of minimum order.
\end{abstract}

\section{Introduction}\label{sec:intro}
Planning for systems with observation and motion uncertainty is generally formulated in the framework of a Partially Observed Markov Decision Process (POMDP), the general solution of which is provided by the Hamilton-Jacobi-Bellman equations \cite{Kumar-book-86}. Attempts to utilize this framework run into the intractability of the computations, referred to as the curse of dimensionality.

In this paper, we provide a structure under which the stochastic optimal control problem can be solved quantifiably near-optimal for moderate levels of noise. We utilize the Wentzell-Freidlin theory of large  deviations for analyzing the asymptotics under small noise \cite{wentzell2012limit}. In particular, we consider a general nonlinear process and measurement models with additive white noise, and compensate the system with feedback. We show that the first-order stochastic error of the stochastic cost function for the feedback-compensated system is distributed according to a Gaussian distribution with zero expected value.

As a result of the independence of the first-order expected error from the feedback law, the optimal zeroth-order (nominal open-loop) control sequence can be designed separately from the optimal closed-loop feedback law; a result which we term as a ``separation of the open-loop and closed-loop designs''. This leads to a novel design approach for partially-observed nonlinear stochastic systems whose characteristics we quantify. We also provide a tractable example of a robotic motion and path planning design based on this theory. Other than the HJB equations, this is the only structure to-date that provides quantifiably near-optimal solutions for a relatively general stochastic optimal control problem. In addition, unlike the HJB, this approach does not run into the problem of curse of dimensionality, as the entire computation is of the order of $ O(Kn^3) $, where $ K $ is the planning horizon, and $ n $ is the state dimension. Lastly, it is observed in simulations that the design is valid for a moderate-range of noise level due to the power of feedback compensation.
\vspace{-5pt}
\section{General Problem}\label{sec:General Problem (Partially Observed System)}\vspace{-2pt}
The general belief space planning problem is formulated as a stochastic control problem in the space of feedback policies. In this section, we define the basic elements of the problem, including system equations and belief dynamics.

\emph{SDE models:} We consider continuous-time Stochastic Differential Equation (SDE) models of the process and measurement as follows:
\begin{subequations}\label{eq:SDEs}
\begin{align}
d\mathbf{x}_{t}&=\mathbf{f}(\mathbf{x}_{t},\mathbf{u}_t)dt+\epsilon\boldsymbol{\sigma}(t)d\boldsymbol{\omega}_{t},\label{eq: continuous-time process model}
\\d\mathbf{z}_{t}&=\mathbf{h}(\mathbf{x}_{t})dt+\epsilon d\boldsymbol{\nu}_{t},\label{eq: continuous-time observation model}
\end{align}
\end{subequations}
where $ \{\boldsymbol{\omega}_{t}, \boldsymbol{\nu}_t, t\ge 0\} $ are two independent standard Wiener processes, $\mathbf{x}\in \mathbb{X} \subset\mathbb{R}^{n_x}$, $\mathbf{u}\in \mathbb{U}\subset\mathbb{R}^{n_u}$, and $\mathbf{z}\in \mathbb{Z}\subset\mathbb{R}^{n_z}$, denote the state, control and observation vectors, respectively, and $\mathbf{f}:\mathbb{X}\times\mathbb{U}\rightarrow\mathbb{X}$, $\mathbf{h}:\mathbb{X}\rightarrow\mathbb{Z}$, $ \boldsymbol{\sigma},\mathbf{a}:\mathbb{R}\rightarrow\mathbb{R}^{n_x\times n_x} $, $ \mathbf{f}=(f_{i})_{0\le i\le n_x}, \mathbf{h}=(h_{i})_{0\le i\le n_z}$, and $ \mathbf{a}:=\boldsymbol{\sigma}\boldsymbol{\sigma}^{T}=(a_{i,j})_{0\le i,j\le n_x} $. We assume that the drift and diffusion coefficients, $ f_{i}, h_i, a_{i,j} $, are bounded and uniformly Lipschitz continuous functions, and the diffusion matrix is uniformly positive-definite. Lastly, $ \mathbf{x}_0\sim \mathcal{N}(\bar{\mathbf{x}}_0, \epsilon^{2}\boldsymbol{\Sigma}_{\mathbf{x}_0}), \epsilon>0 $.

\textit{Belief:} The conditional distribution of the state given the past observations, controls and the initial distribution is termed as ``belief''. In the sequel, we denote the Gaussian belief by $ \mathbf{b}_{t}=(\hat{\mathbf{x}}^{T}_{t}, \mbox{vec}(\mathbf{P}_{t})^{T})^{T}\in \mathbb{B} $, a vector of the mean and covariance of the estimation at time $ t $. 

\begin{problem}\label{problem:Stochastic Control Problem partially observed belief form} \textup{\textbf{Stochastic Control Problem:}} Given an initial belief state $ \mathbf{b}_{0} $, the stochastic optimal control problem is:
\vspace{-1pt}\begin{align}\label{problem eq:Stochastic Control Problem  partially observed belief form}
\nonumber \min_{\boldsymbol{\pi}}~\mathbb{E}[\sum_{t=0}^{K-1}&c_t^{\boldsymbol{\pi}}(\mathbf{b}_t,\mathbf{u}_t)+c_K^{\boldsymbol{\pi}}(\mathbf{b}_K)]
\\ s.t.~\mathbf{b}_{t+1}&=\boldsymbol{\tau}(\mathbf{b}_{t},\mathbf{u}_t,\mathbf{z}_{t+1}),
\end{align}
where the optimization is over Markov policies, $ \mathbbl{\Pi} $, and:
\begin{itemize} 
\item $ J^{\boldsymbol{\pi}}:\mathbbl{\Pi}\rightarrow\mathbb{R} $ is the cost function given the policy $ \boldsymbol{\pi}\in\mathbbl{\Pi} $, and $ J^{\boldsymbol{\pi}}:=\sum_{t=0}^{K-1}c_t^{\boldsymbol{\pi}}(\mathbf{b}_t,\mathbf{u}_t)+c_K^{\boldsymbol{\pi}}(\mathbf{b}_K) $;
\item $ \boldsymbol{\pi}:=\{\boldsymbol{\pi}_{0}, \cdots, \boldsymbol{\pi}_{t}\} $, $ \boldsymbol{\pi}_{t} :\mathbb{B}\rightarrow \mathbb{U} $ and $ \mathbf{u}_{t}=\boldsymbol{\pi}_{t}(\mathbf{b}_{t}) $;
\item $ c^{\boldsymbol{\pi}}_t(\cdot,\cdot):\mathbb{B}\times\mathbb{U}\rightarrow\mathbb{R} $ is the one-step cost function;
\item $ c_K^{\boldsymbol{\pi}}(\cdot):\mathbb{B}\rightarrow\mathbb{R} $  denotes the terminal cost; and
\item $ K\!>\!0 $ is planning horizon, and $ \boldsymbol{\tau} $ defines belief evolution.
\end{itemize}
\end{problem}
\vspace{-3pt}
\section{Method and Main Results}
\textit{Feedback law:} We assume a Lipschitz continuous, bounded and smooth feedback law:
\begin{align}\label{eq:pi feedback law}
\mathbf{u}_{t}=\boldsymbol{\pi}_{t}(\hat{\mathbf{x}}_t).
\end{align}

\textit{Nominal ODEs:} Nominal (unperturbed) trajectories of the system can be obtained using a nominal control sequence (which is calculated using the separation result of this paper). The following Ordinary Differential Equations (ODEs) describe the nominal trajectories:
\begin{align}
\mathbf{\dot{x}}^{p}_{t}=\mathbf{f}(\mathbf{x}^{p}_{t},\mathbf{u}^{p}_t),~
\mathbf{\dot{z}}^{p}_{t}=\mathbf{h}(\mathbf{x}^{p}_{t}),~ \mathbf{u}^{p}_{t}=\boldsymbol{\pi}_{t}(\hat{\mathbf{x}}^{p}_t),
\end{align}
where $ \mathbf{\hat{x}}^{p}_{0}\!:=\!\mathbf{x}^{p}_{0}\!:=\!\mathbb{E}[\mathbf{b}_0] $, and $ \mathbf{x}^{p}_{t} $ is the mean of nominal belief.

\textit{Linearized equations:} We linearize the SDEs of \eqref{eq:SDEs} around nominal trajectories. Thus, if $ |\!|\hat{\mathbf{x}}_t\!-\! \hat{\mathbf{x}}^{p}_t|\!|\!\!\le\!\! \delta $ and $ |\!|\mathbf{x}_t\!-\! \mathbf{x}^{p}_t|\!|\!\!\le\!\! \delta $,
\begin{subequations}\label{eq:linearized system fully observed}
\begin{align}
\mathbf{u}_{t}&\!=\!\mathbf{u}^{p}_{t}-\mathbf{L}_t(\hat{\mathbf{x}}_t\!-\! \hat{\mathbf{x}}^{p}_t)+o(\delta),
\\\mathbf{\dot{x}}_{t}&\!=\!\mathbf{\dot{x}}^{p}_{t}\!+\! \mathbf{A}_t(\mathbf{x}_t \!-\! \mathbf{x}^{p}_{t})\!+\! \mathbf{B}_t(\mathbf{u}_t\!-\!\mathbf{u}^{p}_t) \!+\!\epsilon\mathbf{G}_t\frac{d\boldsymbol{\omega}_t}{dt}\!+\!o(\delta)
\\\nonumber&\!=\!\mathbf{\dot{x}}^{p}_{t}\!+\! \mathbf{A}_t(\mathbf{x}_t \!-\! \mathbf{x}^{p}_{t})\!-\!\mathbf{B}_t\mathbf{L}_t(\hat{\mathbf{x}}_t\!-\! \hat{\mathbf{x}}^{p}_t) \!+\!\epsilon\mathbf{G}_t\frac{d\boldsymbol{\omega}_t}{dt}\!+\!o(\delta),\!\!
\\\mathbf{\dot{z}}_{t}&\!=\!\mathbf{\dot{x}}^{p}_{t}\!+\! \mathbf{H}_t(\mathbf{x}_t \!-\! \mathbf{x}^{p}_{t})\!+\!\epsilon \frac{d\boldsymbol{\nu}_t}{dt}\!+\!o(\delta).
\end{align}
\end{subequations}
with Jacobians (the superscript $ p $ was dropped for simplicity):
\begin{align*}
\mathbf{A}^{p}_t\!:&=\!\nabla_{\mathbf{x}} \mathbf{f}(\mathbf{x},\mathbf{u})|_{ \mathbf{x}^{p}_{t}, \mathbf{u}^{p}_{t}},
\mathbf{B}^{p}_t\!:=\!\nabla_{\mathbf{u}} \mathbf{f}(\mathbf{x},\mathbf{u})|_{ \mathbf{x}^{p}_{t}, \mathbf{u}^{p}_{t}},
\mathbf{G}_t\!:=\!\boldsymbol{\sigma}(t),
\\\mathbf{L}^{p}_t\!:&=\!-\nabla_{\mathbf{x}} \boldsymbol{\pi}_t(\mathbf{x})|_{ \hat{\mathbf{x}}^{p}_{t}},~~~\!
\mathbf{H}^{p}_t\!:=\!\nabla_{\mathbf{x}} \mathbf{h}(\mathbf{x})|_{ \mathbf{x}^{p}_{t}}.
\end{align*}

\textit{Kalman-Bucy Filter (KBF):} The linearized system's estimates can be obtained using the KBF equations:
\begin{subequations}\label{eq:KBF linearized system}
\begin{align}
&\mathbf{\dot{\hat{x}}}_{t} \!\!=\!\! \mathbf{\dot{x}}^{p}_{t}\!\!+\!\! \mathbf{A}_t(\hat{\mathbf{x}}_t \!\!-\!\! \mathbf{x}^{p}_{t})\!\!+\!\! \mathbf{B}_t(\mathbf{u}_t\!\!-\!\! \mathbf{u}^{p}_t) 
\!\!+\!\! \mathbf{K}_{t}(\mathbf{\dot{z}}_{t}\!\!-\!\!\mathbf{\dot{z}}^{p}_{t}\!\!-\!\!\mathbf{H}_{t}(\hat{\mathbf{x}}_t \!\!-\!\! \mathbf{x}^{p}_{t})\!)\!,\!\!\!
\\&\mathbf{\dot{P}}_{t}=\mathbf{A}_{t}\mathbf{P}_{t}+\mathbf{P}_{t}\mathbf{A}_{t}^T+\epsilon^{2}\mathbf{G}_{t}\boldsymbol{\Sigma}_{\boldsymbol{\omega}}\mathbf{G}^{T}_{t}-\epsilon^{2}\mathbf{K}_t\boldsymbol{\Sigma}_{\boldsymbol{\nu}}\mathbf{K}^{T}_t,
\\&\mathbf{K}_t=\epsilon^{-2}\mathbf{P}_{t}\mathbf{H}_{t}^T\boldsymbol{\Sigma}_{\boldsymbol{\nu}}^{-1}.
\end{align}
\end{subequations}
with $ \mathbf{P}_{0}\!:=\!\epsilon^{2}\boldsymbol{\Sigma}_{\mathbf{x}_{0}} $ and $ \mathbf{\hat{x}}^{p}_{0} \!=\! \mathbf{x}^{p}_{0} $, which implies $ \mathbf{\hat{x}}^{p}_{t} \!\equiv\! \mathbf{x}^{p}_{t}, t\ge0 $. 

\textit{Stochastic differential equation governing the evolution of the augmented state:} Since the evolution of the covariance is deterministic, we define $ \mathbf{y}_{t}:=(\mathbf{x}^{T}_{t}, \hat{\mathbf{x}}^{T}_{t})^{T} $ (also denoted by $ \mathbf{y}^{\epsilon}_{t} $), which is the concatenation of the two vectors of state and mean of the belief, and define $ \boldsymbol{\zeta}_{t}:=(\boldsymbol{\omega}^{T}_{t},\boldsymbol{\nu}^{T}_{t})^{T} $. Then, the evolution of this augmented state random variable is:
\begin{align}\label{eq:continuous-time SDE y}
d\mathbf{y}_{t}=\mathbf{g}(t, \mathbf{y}_{t})dt+\epsilon\boldsymbol{\sigma}^{\mathbf{y}}(t)d\boldsymbol{\zeta}_{t},
\end{align}
with $ \mathbf{y}_{0}=(\mathbf{x}_0^{T}, (\mathbf{x}^{p}_0)^{T})^{T} $, where functions $ \mathbf{g}\!:\!\mathbb{R}\!\times\!\mathbb{R}^{2n_x}\!\rightarrow\!\mathbb{R}^{n_x} $ and $ \boldsymbol{\sigma}^{\mathbf{y}}:\mathbb{R}\rightarrow\mathbb{R}^{2n_x\times2n_x} $ are defined (with some abuse of notation) as:
\begin{align*}
&\mathbf{g}(t, \mathbf{y}_{t})\!\!:=\!\!\begin{pmatrix}
\mathbf{f}(\mathbf{x}_{t},\boldsymbol{\pi}_{t}(\hat{\mathbf{x}}_t))\\
\!\!\!\Big(\!\mathbf{f}(\mathbf{x}^{p}_{t},\boldsymbol{\pi}_{t}(\hat{\mathbf{x}}^{p}_t)) \!\!+\!\! \mathbf{A}_t(\hat{\mathbf{x}}_t \!\!-\!\! \mathbf{x}^{p}_{t})
\!\!+\!\! \mathbf{B}_t(\boldsymbol{\pi}_{t}(\hat{\mathbf{x}}_t) \!\!-\!\! \boldsymbol{\pi}_{t}(\hat{\mathbf{x}}^{p}_t) )\!\! \!\\+ \mathbf{K}_{t}(\mathbf{h}(\mathbf{x}_{t})-\mathbf{h}(\mathbf{x}^{p}_{t})-\mathbf{H}_{t}(\hat{\mathbf{x}}_t- \mathbf{x}^{p}_{t}))\!\Big)
\end{pmatrix}\!\!, 
\\&\boldsymbol{\sigma}^{\mathbf{y}}(t):=\begin{pmatrix}
\boldsymbol{\sigma}(t) &\mathbf{0}\\
\mathbf{0} &\mathbf{K}_{t}
\end{pmatrix}.
\end{align*}

\begin{lemma}[Initial State]\label{lemma:initial state} Let $ \{\mathbf{y}^{p}_{t}, t\!\ge\! 0\} $, $ \{\mathbf{y}^{n}_{t}, \!t\ge\! 0 \} $, and
\begin{subequations}
\begin{align}
&\mathbf{\dot{y}}^{p}_{t}=\mathbf{g}(t, \mathbf{y}^{p}_{t}), &&\mathbf{y}^{p}_{0} = ((\mathbf{x}^{p}_0)^{T}, (\mathbf{x}^{p}_0)^{T})^{T},\label{eq: unperturbed ODE}
\\&\mathbf{\dot{y}}^{n}_{t}=\mathbf{g}(t, \mathbf{y}^{n}_{t}), &&\mathbf{y}^{n}_{0} = (\mathbf{x}_0^{T}, \mathbf{x}_0^{T})^{T}.
\end{align}
\end{subequations}
Also, let $ K^{\mathbf{f}} $ and $ K^{\boldsymbol{\pi}_{t}} $ be the Lipschitz constants of $ \mathbf{f} $ and $ \boldsymbol{\pi}_{t} $,
\begin{align*}
c^{\delta}\!\!:=\!\!\frac{\delta}{2}\exp(-\!\!\int_{0}^{K}\!\!\!\!\!K^{\mathbf{f}}(1\!\!+\!\!K^{\boldsymbol{\pi}_{r}})dr),
P_{\delta, \epsilon}\!\!:=\!\!\int_{|\!|\mathbf{x}|\!|\le c^{\delta}}\!\!\!\!\!\!\!\!\!\!\!\!\!\exp(\epsilon^{2}\mathbf{x}^{T}\boldsymbol{\Sigma}_{\mathbf{x}_0}\mathbf{x})d\mathbf{x},
\end{align*}
and $ \delta>0 $. Then,
\begin{align}\label{eq:x_n - x_p nominal}
P\{|\!|(\mathbf{x}^{n}_{K}-\mathbf{x}^{p}_{K})|\!|\le \delta/2\}\ge P_{\delta, \epsilon}.
\end{align}
\end{lemma}

\textit{Linearization of the SDE:} Given $ \mathbf{F}^{\mathbf{g}}_{t}=\nabla_{\mathbf{y}} \mathbf{g}(t,\mathbf{y})|_{t, \mathbf{y}^{p}_{t}} $, we linearize the SDE \eqref{eq:continuous-time SDE y} around ODE \eqref{eq: unperturbed ODE}:
\begin{equation*} 
d\mathbf{{y}}_{t}\!\!=\!\!\mathbf{g}(t,\mathbf{y}^{p}_{t})dt\!+\!\mathbf{F}^{g}_{t}(\mathbf{y}_{t}\!-\!\mathbf{y}^{p}_{t})dt+\! \epsilon\boldsymbol{\sigma}^{\mathbf{y}}(t) d\boldsymbol{w}_{t}+o(|\!|\mathbf{y}_{t}\!-\!\mathbf{y}^{p}_{t}|\!|dt).
\end{equation*}
If $ |\!|\mathbf{y}_{t}-\mathbf{y}^{p}_{t}|\!|\le 2\delta $ (whose asymptotics are calculated using the Wentzell-Freidlin theory, next and Lemma \ref{lemma:initial state}),
\begin{equation}\label{eq:continuous-time SDE y linearized with delta}
d\mathbf{{y}}_{t}\!=\!\mathbf{g}(t,\mathbf{y}^{p}_{t})dt\!+\!\mathbf{F}^{g}_{t}(\mathbf{y}_{t}-\mathbf{y}^{p}_{t})dt+\! \epsilon\boldsymbol{\sigma}^{\mathbf{y}}(t) d\boldsymbol{w}_{t}+o(\delta dt).
\end{equation}

\textit{Action functional \cite{wentzell2012limit}:} For $ [T_1, T_2]\subseteq[0, K] $, the normalized action functional for the family of $ \epsilon $-dependent stochastic processes of \eqref{eq:continuous-time SDE y} is defined as:
\begin{align}\label{eq:action functional time-varying}
S_{T_1,T_2}(\boldsymbol{\phi}):=\frac{1}{2\epsilon^{2}}\int_{T_1}^{T_2}
L(s, \boldsymbol{\phi}_s, \boldsymbol{\dot{\phi}}_{s})ds,
\end{align}
for absolutely continuous $ \boldsymbol{\phi} $, and is set to $ +\infty $ for other $ \boldsymbol{\phi}\in \mathbb{C}_{0K}(\mathbb{R}^{n_{x}}) $ (the space of continuous functions over $ [0, K] $), where $ L\!\!:\!\!\mathbb{R}\!\!\times\!\!\mathbb{R}^{n_x}\!\!\times\!\!\mathbb{R}^{n_x}\!\!\rightarrow\!\!\mathbb{R} $ is the Legendre transform of the cumulant of stochastic process of \eqref{eq:continuous-time SDE y} (assuming $ \mathbf{K}_t\mathbf{K}_t^{T}\!\!\succ\!\!0 $):
\vspace{-1pt}\begin{align}
L(t, \mathbf{x}, \boldsymbol{\beta})=\dfrac{1}{2}(\boldsymbol{\beta}-\mathbf{b}(t,\mathbf{x}))^{T}\mathbf{a}(t, \mathbf{x})^{-1}(\boldsymbol{\beta}-\mathbf{b}(t,\mathbf{x})).
\end{align}

\begin{theorem}[Exponential Rate of Convergence]\label{theorem: Rate of Convergence } Let:
\begin{itemize}
\item $ \mathbb{D} $ be a domain in $ \mathbb{R}^{2n_x} $, and denote its closure by $ \mathrm{cl}(\mathbb{D}) $;
\item $ \partial \mathbb{D} $ denote the boundary of $ \mathbb{D} $;
\item $
\mathbb{H}_{\mathbb{D}}(t, \mathbf{y}^{n}_{0})\!\!=\!\!\{\boldsymbol{\phi}\in\mathbb{C}_{0K}(\mathbb{R}^{n_x}):\boldsymbol{\phi}_{0}=\mathbf{y}^{n}_{0},\boldsymbol{\phi}_{t}\in\mathbb{D}\cup\partial\mathbb{D} \}$.
\end{itemize} 
Assume $ \partial\mathbb{D} = \partial\mathrm{cl}(\mathbb{D}) $. Then, we have the following:
\begin{align}
\lim\limits_{\epsilon\rightarrow 0}\epsilon^{2} \ln P_{\mathbf{y}^{n}_{0}}\{\mathbf{y}^{\epsilon}_{t}\in\mathbb{D} \}&\!\!=\!\!-\!\!\!\!\!\inf\limits_{\boldsymbol{\phi}\in\mathbb{H}_{\mathbb{D}}(t, \mathbf{y}^{n}_{0})}S_{0t}(\boldsymbol{\phi}), \label{eq:asymptotics D_t 1 }
\end{align}
\end{theorem}

\begin{theorem}[Asymptotics of the Diffusion Process]\label{theorem: Asymptotics of the Diffusion Process} Let:
\begin{itemize}
\item $ \mathbb{D}_{t}=\mathrm{cl}(\mathbb{B}^{c}_{\delta/2}(\mathbf{y}^{n}_{t})) $, the closure of the complement of a ball with radius $ \delta/2>0 $ around the point $ \mathbf{y}^{n}_{t} $; and
\item $ \tau^{\epsilon}=\mathrm{Min}\{t:\mathbf{y}^{\epsilon}_{t}\in\mathbb{D}_{t} \} $.
\end{itemize} Then, 
\begin{align}
\lim_{\epsilon \to 0} \epsilon^2 \ln P_{\mathbf{y}^{n}_0}\{\tau^\epsilon \leq t\} 
\!=\! -\!\!\!\!\!\!\! \inf_{\{\boldsymbol{\phi}: \boldsymbol{\phi}_0 = \mathbf{y}^{n}_0, |\!|\boldsymbol{\phi}_t - \mathbf{y}_t^n|\!| > \delta/2 \}} S_{0t}(\boldsymbol{\phi}).
\end{align}
\end{theorem}
Proofs of Theorems \ref{theorem: Rate of Convergence } and \ref{theorem: Asymptotics of the Diffusion Process} can be found in \cite{wentzell2012limit}.

\textit{Nominal belief:} Starting from $ \mathbf{b}^{p}_{0}= \mathbf{b}_{0} $, the nominal belief evolution is given by $ \mathbf{\dot{b}}^{p}_{t}=\boldsymbol{\tau}(\mathbf{b}^{p}_{t},\mathbf{u}^{p}_t,\mathbf{\dot{z}}^{p}_{t}) $. Given equations \eqref{eq:KBF linearized system}, $ \mathbf{b}^{p}_{t}=((\hat{\mathbf{x}}^{p}_{t})^{T}, \mbox{vec}(\mathbf{P}^{p}_{t})^{T})^{T}
= ((\mathbf{x}^{p}_{t})^{T}, \mbox{vec}(\mathbf{P}_{t})^{T})^{T} $, and linearizing $ \boldsymbol{\tau} $ only involves linearization of mean evolution:
\begin{align*}
\mathbf{\dot{b}}_{t}\!\!=\!\! \mathbf{\dot{b}}^{p}_{t}\!+\! \mathbf{T}^{\mathbf{b}}_t(\mathbf{b}_t\!-\!\mathbf{b}^{p}_t)\!+\!\mathbf{T}^{\mathbf{u}}_t(\mathbf{u}_t\!-\!\mathbf{u}^{p}_t)\!+\! \mathbf{T}^{\mathbf{z}}_t(\mathbf{\dot{z}}_{t}\!-\!\mathbf{\dot{z}}^{p}_{t})\!+\!o(\delta),\!\!\!\!
\end{align*}
with the Jacobians defined as usual.

\textit{Linearization of belief and cost:} To address problem \eqref{problem:Stochastic Control Problem partially observed belief form}, we discretize the equations \eqref{eq:linearized system fully observed} in time with the discretization interval of $ dt \equiv 1 $. Let
$ J^{p}:=\sum_{t=0}^{K-1}c_t(\mathbf{b}^{p}_t,\mathbf{u}^{p}_t)+c_K(\mathbf{b}^{p}_K) $, and linearize the cost function $ J^{\boldsymbol{\pi}} $ around the nominal trajectories:
\begin{align}
J^{\boldsymbol{\pi}}=J^{p}+\tilde{J}_1+o(\delta),
\end{align}
with $
\tilde{J}_1= \sum_{t=0}^{K-1} (\mathbf{C}^{\mathbf{b}}_t(\mathbf{b}_t-\mathbf{b}^{p}_t)+ \mathbf{C}^{\mathbf{u}}_t(\mathbf{u}_t-\mathbf{u}^{p}_t))+ \mathbf{C}^{\mathbf{b}}_K(\mathbf{b}_K-\mathbf{b}^{p}_K) $, where the Jacobians are defined as usual.

If $ |\!|\mathbf{x}_{K}-\mathbf{x}^{n}_{K}|\!|\le \delta/2 $ and $ |\!|\mathbf{x}^{n}_{K}-\mathbf{x}^{p}_{K}|\!|\le \delta/2 $, using the triangle inequality (note: as $ \epsilon\downarrow 0 $, using Theorems \ref{theorem: Rate of Convergence }, \ref{theorem: Asymptotics of the Diffusion Process}, and Lemma \ref{lemma:initial state}, the probability of the first and second events tend exponentially to one, respectively; similarly for $ |\!|\hat{\mathbf{x}}_{K}-\mathbf{x}^{p}_{K}|\!| $):
\begin{align}
|\!|\mathbf{x}_{K}-\mathbf{x}^{p}_{K}|\!|\le& |\!|\mathbf{x}_{K}-\mathbf{x}^{n}_{K}|\!|+|\!|\mathbf{x}^{n}_{K}-\mathbf{x}^{p}_{K}|\!|
\le\delta,
\end{align}
which means that all the linearizations are valid with a probability that tends to one as $ \epsilon\downarrow0 $.

\begin{theorem}[First Order Cost Function Error]\label{theroem:First Order Cost Function Error (Fully Observed Case)} For a time-discrete system, under a first-order approximation for the small noise paradigm, the stochastic cost function is dominated by the nominal part of the cost function, and the expected first-order error is zero:
\begin{align*}
\mathbb{E}[\tilde{J}_{1}]=0.
\end{align*}
Moreover, if the initial, process, and observation noises at each time are distributed according to zero mean Gaussian distributions, then $ \tilde{J}_{1} $ also has a zero mean Gaussian distribution.
\end{theorem}

\begin{coroll}\textit{Separation of the Open-Loop and Closed-Loop Designs Under Small Noise:}\label{coroll 1:Cost first order fully observed}
Based on Theorem \ref{theroem:First Order Cost Function Error (Fully Observed Case)}, under the small noise paradigm, as $ \epsilon\downarrow0 $, the design of the feedback law can be conducted separately from the design of the open loop optimized trajectory. Furthermore, this result holds with a probability that exponentially tends to one as $ \epsilon\downarrow0 $.
\end{coroll}

Our separation principle combined with the usual separation principle provides a design structure where the optimal designs of the control law, nominal trajectory and estimator can be separated from each other. Thus, we couple the latter two, and design a nominal trajectory that aims for the best nominal estimation performance, which coincides with the Trajectory-optimized Linear Quadratic Gaussian (T-LQG) design \cite{rafieisakhaei2016belief}. 

\begin{problem}\label{problem:Planning Problem}\textup{\textbf{Trajectory Planning Problem:}} Given an initial belief $ \mathbf{b}_{0} $, a goal region of a ball with radius $ r_g $ around a goal state $ \mathbf{x}_{g}\in\mathbb{X} $, horizon $ K>0 $, and $ \mathbf{W}^{u}_{t}\succeq 0$, solve:
\begin{subequations}
\begin{align}
\nonumber \min_{\mathbf{u}^{p}_{0:K-1}}\sum\limits_{t=1}^{K}[&\mathrm{tr}( \mathbf{P}^{+}_{\mathbf{b}^{p}_t})+({\mathbf{u}}^{p}_{t-1})^{T}\mathbf{W}^{u}_{t}{\mathbf{u}}^{p}_{t-1}]
\\s.t.~~\mathbf{P}^{-}_{t}&=\mathbf{A}_{t-1}\mathbf{P}^{+}_{t-1}\mathbf{A}_{t-1}^T+\epsilon^{2}\mathbf{G}_{t-1}\boldsymbol{\Sigma}_{\boldsymbol{\omega}}\mathbf{G}_{t-1}^T\label{eq:riccati 1},
\\\mathbf{S}_{t}&=\mathbf{H}_t\mathbf{P}^{-}_{t}\mathbf{H}_t^T+\epsilon^{2}\boldsymbol{\Sigma}_{\boldsymbol{\nu}}\label{eq:riccati 2},
\\\mathbf{P}^{+}_{t}&=(\mathbf{I}-\mathbf{P}^{-}_{t}\mathbf{H}_t^T\mathbf{S}_{t}^{-1}\mathbf{H}_t)\mathbf{P}^{-}_{t}\label{eq:riccati 4},
\\\mathbf{P}^{+}_{0}&=\epsilon^{2}\boldsymbol{\Sigma}_{\mathbf{x}_{0}}\label{eq:initial cov},
\\\mathbf{x}^{p}_{0} &= \mathbb{E}[\mathbf{b}_{0}]\label{eq:initial mean},
\\\mathbf{x}^{p}_{t+1}&=\mathbf{f}(\mathbf{x}^{p}_t, \mathbf{u}^{p}_t), ~0\!\le\! t\! \le\! K\!-\!1\label{eq:state propagation},
\\|\!|\mathbf{x}^{p}_{K}-\mathbf{x}_g|\!|_{2}&<r_g\label{eq:terminal constraint},
\\|\!|\mathbf{u}^{p}_{t}|\!|_{2}&\le r_u, ~1\!\le\! t\! \le\! K.\label{eq:control contraint}
\end{align}
\end{subequations}
\end{problem}

\textit{Control policy:} After linearizing the equations around the optimized nominal trajectory, the resulting control policy is a linear feedback policy \cite{Kumar-book-86}, $ \mathbf{u}_{t}=\mathbf{u}^{p}_{t}-\mathbf{L}^{p}_{t}(\hat{\mathbf{x}}_t-\mathbf{x}^{p}_t) $, where the feedback gain $ \mathbf{L}^{p}_{t} $ is:
\begin{align*}
\mathbf{L}^{p}_{t} = (\mathbf{W}^{u}_{t}+(\mathbf{B}^{p}_{t})^T\mathbf{P}^{f}_{t+1}\mathbf{B}^{p}_{t})^{-1}(\mathbf{B}^{p}_{t})^T\mathbf{P}^{f}_{t+1}\mathbf{A}^{p}_{t},
\end{align*}
and the matrix $ \mathbf{P}^{f}_{t} $ is the result of backward iteration of the dynamic Riccati equation
\begin{align*}
&\mathbf{P}^{f}_{t-1} = (\mathbf{A}^{p}_{t})^T\mathbf{P}^{f}_{t}\mathbf{A}^{p}_{t}
\\&-\!(\mathbf{A}^{p}_{t})^T\mathbf{P}^{f}_{t}\mathbf{B}^{p}_{t}(\mathbf{W}^{u}_{t}+(\mathbf{B}^{p}_{t})^T\mathbf{P}^{f}_{t}\mathbf{B}^{p}_{t})^{-1}(\mathbf{B}^{p}_{t})^T\mathbf{P}^{f}_{t}\mathbf{A}^{p}_{t}\!+\!\mathbf{W}^{x}_{t},
\end{align*}
which is solvable with a terminal condition $ \mathbf{P}^{f}_{K}=\mathbf{W}^{x}_{t}\succeq 0 $.

\section{Simulation Results}
\begin{figure}[htb!]
\centering
\includegraphics[width=0.8\linewidth]{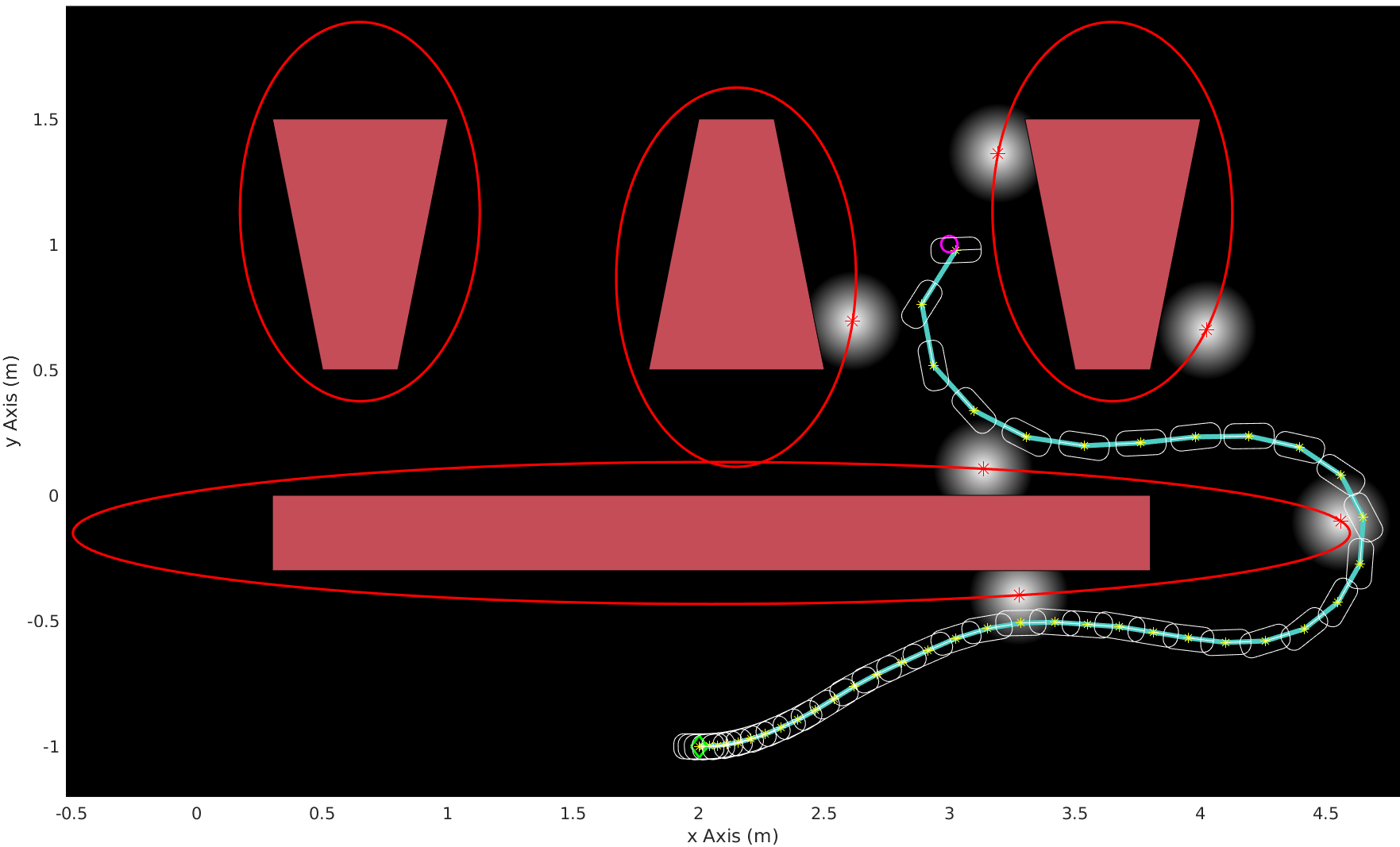}
   \caption{Optimized planned trajectory for a car-like robot. Landmarks are shown with light areas. The planning horizon is 40 steps. Ellipsoids show the safety margin of the collision-avoidance function. Initial state is $ (x, y, \theta)=(2, -1, 0) $, goal state is $ (3, 1, 0) $, $ \boldsymbol{\Sigma}_{\mathbf{x}_0}\!=\!\boldsymbol{\Sigma}_{\boldsymbol{\omega}}\!=\!0.01\mathbf{I}_{n_x} $, and $ \boldsymbol{\Sigma}_{\boldsymbol{\nu}} \!=\! 0.01\mathbf{I}_{n_z} $. \label{fig:plan}}\vspace{-8pt}
\end{figure}

\begin{figure}[htb!]
\centering
\includegraphics[width=0.8\linewidth]{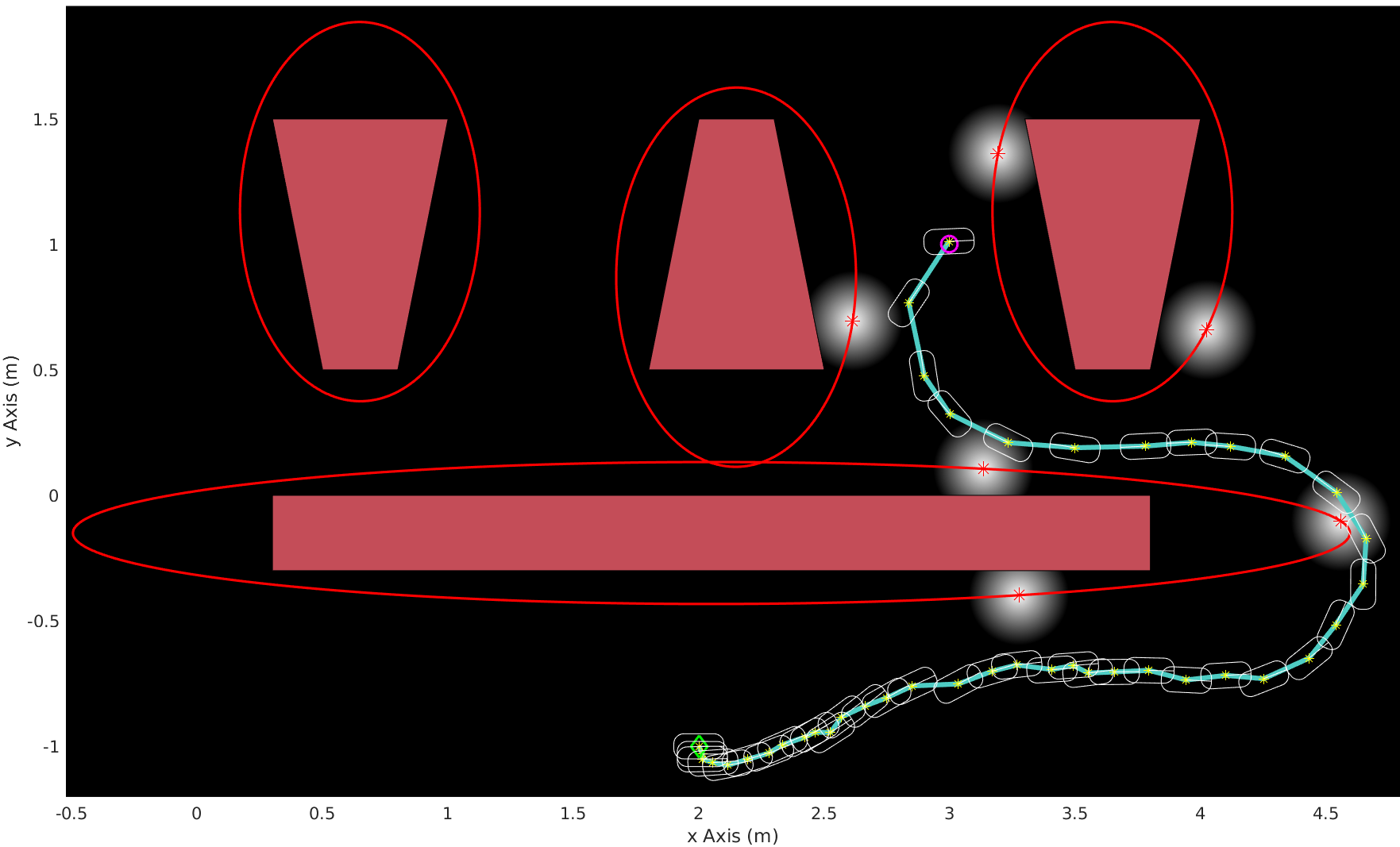}
   \caption{A typical execution trajectory. Since, no significant deviation occurred, planning was only performed once.\label{fig:exec}}\vspace{-5pt}
\end{figure}

\begin{figure}[htb!]
\centering
\includegraphics[width=0.8\linewidth]{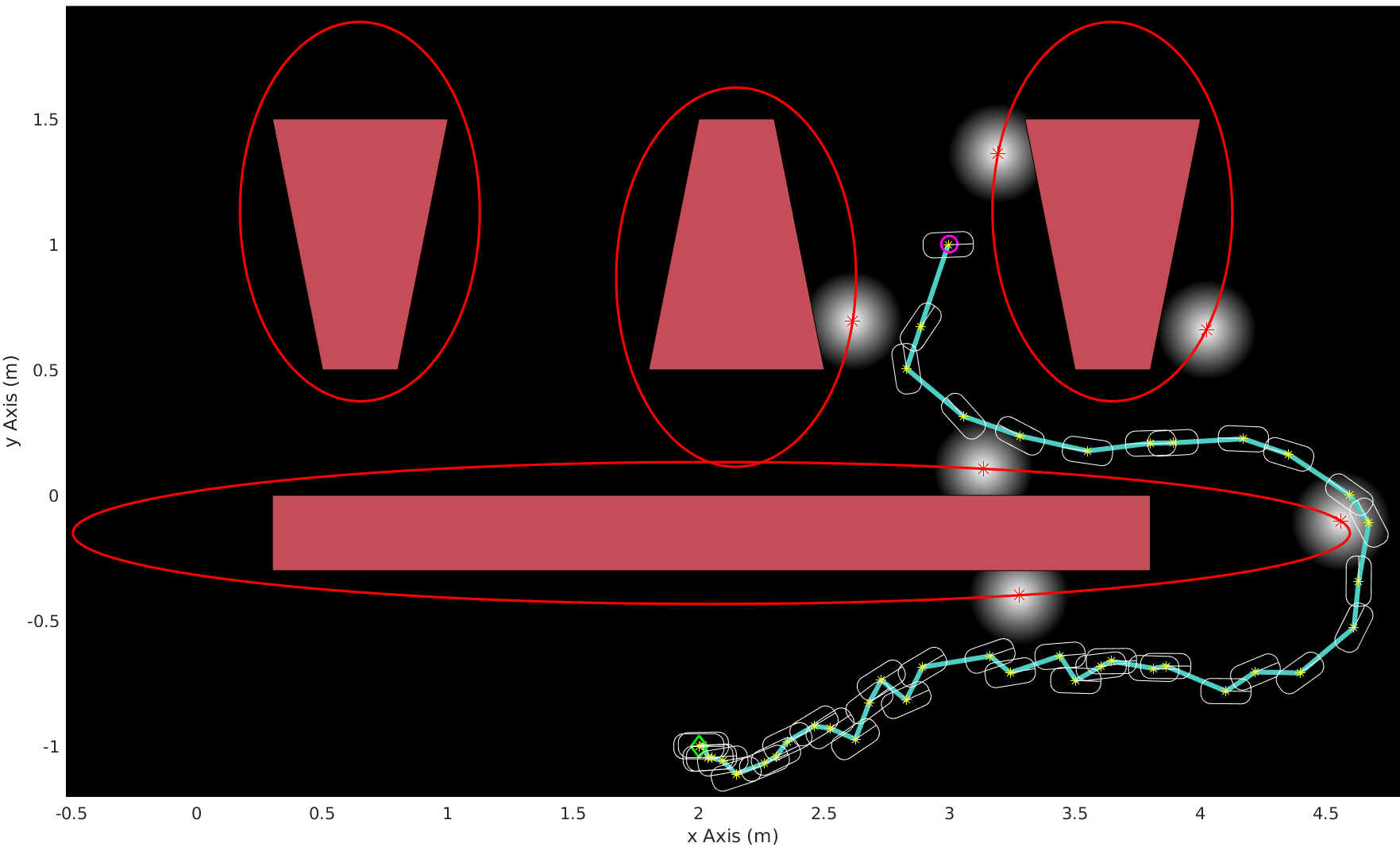}
   \caption{Estimate trajectory. Although KF is used for planning, Extended KF is used during execution for better performance.\label{fig:estimate}} 
\end{figure}

We consider a non-holonomic car-like robot in an environment with road-blocks equipped with landmark-based range and bearing measurement model. We use the MATLAB $ \texttt{fmincon} $ optimizer with no initial trajectory to obtain the solution of problem \eqref{problem:Planning Problem}. For collision-avoidance, we utilize the Obstacle Barrier Function (OBF) method of \cite{rafieisakhaei2016belief}. Figures \ref{fig:plan}, \ref{fig:exec}, and \ref{fig:estimate} show the optimized planned, execution, and estimate trajectories, respectively.

\section{Conclusion}
We considered the general problem of controlling a stochastic nonlinear system with process and measurement uncertainties. We used the Wentzell-Freidlin theory of large deviations and provided a novel result of a ``separation of the open-loop and closed-loop designs''. This result, combined with the usual separation principle (of the estimator and controller designs) leads to an asymptotically-optimal design approach of the Trajectory-optimized Linear Quadratic Gaussian (T-LQG) under small noise, and a near-optimal design for moderate noise levels involving a polynomial order of calculations of minimum order.

\footnotesize\bibliographystyle{IEEEtran}
\bibliography{AliAgha}

\end{document}